\documentclass[letterpaper, 10 pt, conference]{ieeeconf}  

\IEEEoverridecommandlockouts                              
\usepackage{graphicx}
\usepackage{amsmath} 
\usepackage{amssymb}  
\usepackage{booktabs}       
\usepackage{color}
\usepackage{tikz}

\title{\LARGE \bf
Mathematics of Deep Learning
}

\author{Ren\'e Vidal \and Joan Bruna \and Raja Giryes \and Stefano Soatto
\thanks{R. Vidal is with the Center for Imaging Science, Biomedical Engineering, Johns Hopkins University, Baltimore, USA {\tt\small rvidal@cis.jhu.edu}}%
\thanks{J. Bruna is with the Courant Institute of Mathematical Sciences, Center for Data Science, New York University, USA {\tt \small bruna@cims.nyu.edu}}%
\thanks{R. Giryes is with the School of Electrical Engineering, Tel-Aviv University, Tel Aviv, Israel {\tt\small raja@tauex.tau.ac.il}}%
\thanks{S. Soatto is with the Department of Computer Science, University of California, Los Angeles, USA {\tt\small soatto@ucla.edu}}%
}

\def\Re{\mathbb{R}}

\graphicspath{{./figures/}}
\newcommand{\KL}[2]{\operatorname{KL}(\,#1\,\|\,#2\,)}
\begin{document}

\maketitle
\thispagestyle{empty}
\pagestyle{empty}

\begin{abstract}
Recently there has been a dramatic increase in the performance of recognition systems due to the introduction of deep architectures for representation learning and classification. However, the mathematical reasons for this success remain elusive. This tutorial will review recent work that aims to provide a mathematical justification for several properties of deep networks, such as global optimality, geometric stability, and invariance of the learned representations.
\end{abstract}

\section{Introduction}

Deep networks \cite{lecun2015deep} are parametric models that perform sequential operations on their input data. Each such operation, colloquially called a ``layer'', consists of a linear transformation, say, a convolution of its input, followed by a point-wise nonlinear ``activation function'', e.g., a sigmoid. Deep networks have recently led to dramatic improvements in classification performance in various applications in speech and natural language processing, and computer vision. The crucial property of deep networks that is believed to be the root of their performance is that they have a large number of layers as compared to classical neural networks; but there are other architectural modifications such as rectified linear activations (ReLUs) \cite{nair2010rectified} and residual ``shortcut'' connections~\cite{he2016identity}. Other major factors in their success is the availability of massive datasets, say, millions of images in datasets like ImageNet~\cite{Socher2009}, and efficient GPU computing hardware for solving the resultant high-dimensional optimization problem which may have up to $100$ million parameters.

The empirical success of deep learning, especially convolutional neural networks (CNNs) for image-based tasks, presents numerous puzzles to theoreticians. In particular, there are three key factors in deep learning, namely the architectures, regularization techniques and optimization algorithms, which are critical to train well-performing deep networks and understanding their necessity and interplay is essential if we are to unravel the secrets of their success.

\subsection{Approximation, depth, width and invariance properties}

An important property in the design of a neural network architecture is its ability to approximate arbitrary functions of the input. But how does this ability depend on parameters of the architecture, such as its depth and width?  Earlier work shows that neural networks with a single hidden layer and sigmoidal activations are universal function approximators
\cite{Cybenko1989,hornik1989multilayer,Hornik1991,barron1994approximation}. However, the capacity of a wide and shallow network can be replicated by a deep network with significant improvements in performance. One possible explanation is that deeper architectures are able to better capture invariant properties of the data compared to their shallow counterparts. In computer vision, for example, the category of an object is invariant to changes in viewpoint, illumination, etc. While a mathematical analysis of why deep networks are able to capture such invariances remains elusive, recent progress has shed some light on this issue for certain sub-classes of deep networks. In particular, scattering networks \cite{bruna2013invariant} are a class of deep networks whose convolutional filter banks are given by complex, {\bf multi-resolution wavelet families.} As a result of this extra structure, they are provably stable and locally invariant signal representations, and reveal the fundamental role of geometry and stability that underpins the generalization performance of  modern deep convolutional network architectures; see Section \ref{sec:scattering}.

\subsection{Generalization and regularization properties}
Another critical property of a neural network architecture is its ability to generalize from a small number of training examples. Traditional results from statistical learning theory \cite{bartlett2003vapnik} show that the number of training examples needed to achieve good generalization grows polynomially with the size of the network. In practice, however, deep networks are trained with much fewer data than the number of parameters ($N \ll D$ regime) and yet they can be prevented from overfitting using very simple (and seemingly counter-productive) regularization techniques like Dropout~\cite{Srivastava:JMRL2014}, which simply freezes a random subset of the parameters at each iteration.

One possible explanation for this conundrum is that deeper architectures produce an embedding of the input data that approximately preserves the distance between data points in the same class, while increasing the separation between classes. This tutorial will overview the recent work of \cite{Giryes:TSP2016}, which uses tools from {\bf compressed sensing and dictionary learning} to prove that deep networks with random Gaussian weights perform a distance-preserving embedding of the data in which similar inputs are likely to have a similar output. These results provide insights into the metric learning properties of the network and lead to bounds on the generalization error that are informed by the structure of the input data.

\subsection{Information-theoretic properties}

Another key property of a network architecture is its ability to produce a good ``representation of the data''. Roughly speaking, a representation is any function of the input data that is useful for a task. An optimal representation would be one that is ``most useful'' as quantified, for instance, by information-theoretic, complexity or invariance criteria~\cite{tishby2000information}. This is akin to the ``state'' of the system and is what an agent would store in its memory in lieu of the data to predict future observations. For example, the state of a Kalman filter is an optimal representation for predicting data generated by a linear dynamical system with Gaussian noise; in other words, it is a minimal sufficient statistic for prediction. For complex tasks where the data may be corrupted by ``nuisances'' that do not contain information about the task, one may also wish for this representation to be ``invariant'' to such nuisances so as not to affect future predictions. In general, optimal representations for a task can be defined as sufficient statistics (of past or ``training'' data) that are also minimal, and invariant to nuisance variability affecting future (``test'') data~\cite{soattoC16ICLR}. Despite a large interest in representation learning,
a comprehensive theory that explains the performance of deep networks as constructing optimal representations~does~not~yet exist. Indeed, even basic concepts such as sufficiency and in-variance have received diverse treatment~\cite{bruna2013invariant,soattoC16ICLR,poggio}.


Recent work~\cite{shwartz2017opening,achille2016information,alemi2016deep} has begun to establish information-theoretic foundations for representations learned by deep networks. These include the observation that the information bottleneck loss~\cite{tishby2000information}, which defines a relaxed notion of minimal sufficiency, can be used to compute \emph{optimal} representations. The information bottleneck loss can be re-written as the sum of a cross-entropy term, which is precisely the most commonly used loss in deep learning, with an additional regularization term. The latter can be implemented by introducing noise, similar to adaptive dropout noise, in the learned representation~\cite{achille2016information}. The resulting form of regularization, called \emph{information dropout} in~\cite{achille2016information}, shows improved learning under resource constraints, and can be shown to lead to ``maximally disentangled'' representations, i.e., the (total) correlation among components of the representation is minimal, thereby making the features indicators of independent characteristics of data. Moreover, similar techniques show improved robustness to adversarial perturbations~\cite{alemi2016deep}. {\bf Information theory}  is hence expected to play a key role in formalizing and analyzing the properties of deep representations and suggesting new classes of regularizers.

\def\layersep{2.5cm}
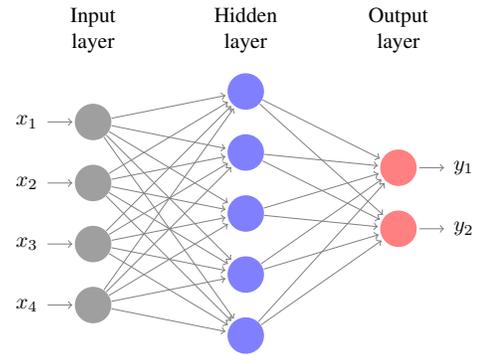
\begin{figure}[t]
\centering
\resizebox{0.75 \columnwidth}{!}{
\begin{tikzpicture}[shorten >=1pt,->,draw=black!50, node distance=\layersep]
    \tikzstyle{every pin edge}=[<-,shorten <=1pt]
    \tikzstyle{neuron}=[circle,fill=black!25,minimum size=17pt,inner sep=0pt]
    \tikzstyle{input neuron}=[neuron, fill=gray!75];
    \tikzstyle{output neuron}=[neuron, fill=red!50];
    \tikzstyle{hidden neuron}=[neuron, fill=blue!50];
    \tikzstyle{annot} = [text width=4em, text centered]

    \foreach \name / \y in {1,...,4}
        \node[input neuron, pin=left:$x_\y$] (I-\name) at (0,-\y) {};

    \foreach \name / \y in {1,...,5}
        \path[yshift=0.5cm]
            node[hidden neuron] (H-\name) at (\layersep,-\y cm) {};

    \foreach \name / \y in {1,...,2}
    \node[output neuron,pin={[pin edge={->}]right:$y_\y$}, right of=H-3] (O-\name) at (\layersep,-0.75-\y) {};

    \foreach \source in {1,...,4}
        \foreach \dest in {1,...,5}
            \path (I-\source) edge (H-\dest);

    \foreach \source in {1,...,5}
        \foreach \dest in {1,...,2}
        \path (H-\source) edge (O-\dest);

    \node[annot,above of=H-1, node distance=1cm] (hl) {Hidden layer};
    \node[annot,left of=hl] {Input layer};
    \node[annot,right of=hl] {Output layer};
\end{tikzpicture}
}
\caption{Illustration of a neural neural network with $D=d_1=4$ inputs, $d_2=5$ hidden layers, and $C=d_3=2$ outputs. Here, the output can be written as $y=(y_1,y_2) =\psi_2(\psi_1(xW^1)W^2)$, where $x = (x_1,\dots,x_4)$ is the input, $W^1\in\Re^{4\times 5}$ is the matrix of weights from the input layer to the hidden layer, $W^2\in\Re^{5\times 2}$ is the matrix of weights from the hidden to the output layer, and $\psi_1$ and $\psi_2$ are activation functions.}
\label{fig:NN-example}
\vspace{-2mm}
\end{figure}

\subsection{Optimization properties}

The classical approach to training neural networks is to minimize a (regularized) loss using backpropagation \cite{werbos1974beyond}, a gradient descent method specialized to neural networks.
Modern versions of backpropagation rely on stochastic gradient descent (SGD) to efficiently approximate the gradient for massive datasets. While SGD has been rigorously analyzed only for convex loss functions~\cite{schmidt2013minimizing}, in deep learning the loss is a non-convex function of the network parameters, hence there are no guarantees that SGD finds the global minimizer.

In practice, there is overwhelming evidence that SGD routinely yields good solutions for deep networks. Recent work on understanding the quality of training 
argues that critical points are more likely to be saddle points rather than spurious local minima \cite{Dauphin:NIPS2014} and that local minima concentrate near the global optimum \cite{Choromanska:AIS2015}. Recent work has also revealed that local minima discovered by SGD that lead to good generalization error belong to very flat regions of the parameter space \cite{dinh2017sharp}. This motivates algorithms like Entropy-SGD that are specialized to find such regions and draw from similar results in the analysis of binary perceptrons in statistical physics~\cite{baldassi2016local}; they have been shown to perform well on deep networks~\cite{chaudhari2016entropy}. Surprisingly, these techniques from statistical physics are intimately connected to the regularization properties of partial differential equations (PDEs)~\cite{chaudhari2017deep}. For instance, local entropy, the loss that Entropy-SGD minimizes, is the solution of the {\bf Hamilton-Jacobi-Bellman PDE} and can therefore be written as a {\bf stochastic optimal control} problem, which penalizes greedy gradient descent. This direction further leads to connections between variants of SGD with good empirical performance and standard methods in convex optimization such as inf-convolutions and proximal methods. Researchers are only now beginning to unravel the loss functions of deep networks in terms of their topology, which dictates the complexity of optimization, and their geometry, which seems to be related to generalization properties of the classifiers \cite{freeman2016topology,Sokolic2016,Sokolic2017b}.

This tutorial will overview recent work showing that the error surface for high-dimensional non-convex optimization problems such as deep learning has some benign properties. For example, the work of \cite{Haeffele:CVPR17,Haeffele:arXiv2015-PosFactor} shows that for certain classes of neural networks for which both the loss function and the regularizer are sums of positively homogeneous functions of the same degree, a local optimum such that many of its entries are zero is also a global optimum. These results will also provide a possible explanation for the success of RELUs, which are positively homogeneous functions. Particular cases of this framework include, in addition to deep learning, matrix factorization~and~tensor~factorization~\cite{Haeffele:ICML14}.

\subsection{Paper outline}
The remainder of this paper is organized as follows. Section \ref{sec:preliminaries} describes the input-output map of a deep network. Section \ref{sec:optimization} studies the problem of training deep networks and establishes conditions for global optimality. Section \ref{sec:scattering} studies invariance and stability properties of scattering networks. Section \ref{sec:structure-generalization} studies structural properties of deep networks, such as metric properties of the embedding as well as bounds on the generalization error. Section \ref{sec:info-theory} studies information-theoretic properties of deep representations.

\section{Preliminaries}
\label{sec:preliminaries}
A deep network is a hierarchical model where each layer applies a linear transformation followed by a non-linearity to the preceding layer. Specifically, let $X \in \Re^{N\times D}$ be the input data,\footnote{For the sake of simplicity, we assume that inputs lie in $\Re^D$, but many of the results in this paper apply also to more general inputs such as tensors (e.g., RGB data), in which case the weights become tensors as well.} where each row of $X$ is a $D$-dimensional data point (e.g., a grayscale image with $D$ pixels) and $N$ is the number of training examples. Let $W^k\in\Re^{d_{k-1}\times d_k}$ be a matrix representing a \emph{linear transformation} applied to the output of layer $k-1$, $X_{k-1}\in\Re^{N\times d_{k-1}}$,  to obtain a $d_k$-dimensional representation $X_{k-1} W^k \in \Re^{N\times d_k}$ at layer $k$. For example, each column of $W^k$ could represent a convolution with some filter (as in convolutional neural networks) or the application of a linear classifier (as in fully connected networks). Let $\psi_k : \Re \to \Re$ be a non-linear \emph{activation function}, e.g., a hyperbolic tangent $\psi_k(x) = \tanh(x)$, a sigmoid $\psi_k(x) = (1+e^{-x})^{-1}$, or a rectified linear unit (ReLU) $\psi_k(x) = \max\{0,x\}$.\footnote{More broadly, $\psi_k$ could be a many to one function, such as max pooling.} This non-linearity is applied to each entry of $X_{k-1} W^k$ to generate the $k$th layer of a neural network as $X_k = \psi_k(X_{k-1} W^k)$. The output $X_K$ of the network is thus given by (see Fig.~\ref{fig:NN-example}):
\begin{equation}
\label{eq:Phi_basic_NN}
\begin{split}
\Phi(X, W^1,&\ldots,W^K) = \psi_K(\psi_{K-1} ( \cdots \\
  &\psi_2(\psi_1(X W^1) W^2) \cdots W^{K-1}) W^K).
\end{split}
\end{equation}
Note that $\Phi$ is an $N\times C$ matrix, where $C = d_K$ is the dimension of the output of the network, which is equal to the number of classes in the case of a classification problem. Notice also that the map $\Phi$ can be seen as a function of the network weights $W=\{W^k\}_{k=1}^K$ with a fixed input $X$, as will be the case when we discuss the training problem in \ref{sec:optimization}. Alternatively, we can view the map $\Phi$ as a function of the input data $X$ with fixed weights $W$, as will be the case when we discuss properties of this map in Sections \ref{sec:scattering}-\ref{sec:info-theory}.

\section{Global Optimality in Deep Learning}
\label{sec:optimization}

This section studies the problem of learning the parameters $W = \{W^k\}_{k=1}^K$ of a deep network from $N$ training examples $(X,Y)$. In a classification setting, each row of $X\in\Re^{N\times D}$ denotes a data point in $\Re^D$ and each row of $Y \in \{0,1\}^{N\times C}$ denotes the membership of each data point to one out of $C$ classes, i.e.,  $Y_{jc} = 1$ if the $j$th row of $X$ belongs to class $c\in\{1,\dots, C\}$ and $Y_{jc} = 0$ otherwise. In a regression setting, the rows of $Y \in \Re^{N\times C}$ denote the dependent variables for the rows of $X$. The problem of learning the network weights $W$ is formulated as the following optimization problem:
\begin{equation}
\!
\min_{\{W^k\}_{k=1}^K} \!\! \ell(Y, \Phi(X,W^1, \dots, W^K)) + \lambda \Theta(W^1, \dots, W^K),\!
\label{eq:non-convex-optimization-problem}
\end{equation}
where $\ell(Y,\Phi)$ is a \emph{loss function} that measures the agreement between the true output, $Y$, and the predicted output, $\Phi(X,W)$ in \eqref{eq:Phi_basic_NN}, $\Theta$ is a \emph{regularization function}  designed to prevent overfitting, e.g., weight decay via $\ell_2$ regularization $\Theta(W) \!=\! \sum_{k=1}^K \|W^k\|_F^2$,  and $\lambda \!>\! 0$ is a balancing parameter.

\subsection{The challenge of non-convexity in neural network training}
\begin{figure}
\includegraphics[width=\linewidth,clip=true,trim=0 2 425 34]{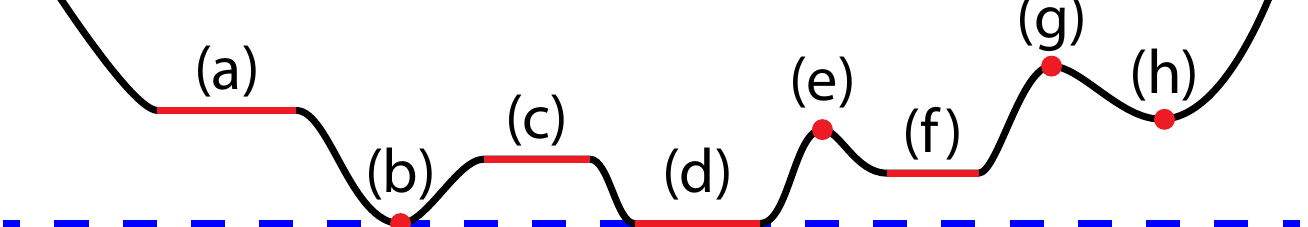}
\caption{Example critical points of a non-convex function (shown in red). (a,c) Plateaus. (b,d) Global minima. (e,g) Local maxima. (f,h) Local minima.}
\label{fig:local_min}
\vspace{-4mm}
\end{figure}

A key challenge in neural network training is that the optimization problem in \eqref{eq:non-convex-optimization-problem} is \emph{non-convex} because, even if the loss $\ell(Y,\Phi)$ is typically a convex function of $\Phi$, e.g., the squared loss $\ell(Y,\Phi) = \|Y - \Phi\|_F^2$, the map $\Phi(X,W)$ is usually a non-convex function of $W$ due to the product of the $W^k$ variables and the nonlinearities $\psi_k$ in \eqref{eq:Phi_basic_NN}. This presents significant challenges to existing optimization algorithms -- \\ including (but certainly not limited to) gradient descent, stochastic gradient descent, alternating minimization, block coordinate descent, back-propagation, and quasi-Newton methods -- which are typically only guaranteed to converge to a critical point of the objective function~\cite{Mairal:JMLR2010,Rumelhart:CogModel1988,Wright:1999,Xu:SIAM2013}. However, for non-convex problems, the set of critical points includes not only global minima, but also local minima, local maxima, saddle points and saddle plateaus, as  illustrated in Fig.~\ref{fig:local_min}. As a result, the non-convexity of the problem leaves the model somewhat ill-posed in the sense that it is not just the model formulation that is important but also implementation details, such as how the model is initialized and particulars of the optimization algorithm, which can have a significant impact on the performance of the model.

To address the issue of non-convexity, a common strategy used in deep learning is to initialize the network weights at random, update these weights using local descent, check if the training error decreases sufficiently fast, and if not, choose another initialization. In practice, it has been observed that if the size of the network is large enough, this strategy can lead to markedly different solutions for the network weights, which give nearly the same objective values and classification performance \cite{Choromanska:AIS2015}. It has also been observed that when the size of the network is large enough and the non-linearity is chosen to be a ReLU, many weights are zero, a phenomenon known as \emph{dead neurons}, and the classification performance significantly improves  \cite{Dahl:ICASSP2013,Maas:ICML2013, Krizhevsky:NIPS2012,Zeiler:ICASSP2013}. While this empirically suggests that when the size of the network is large enough and ReLU non-linearities are used \emph{all local minima could be global}, there is currently no rigorous theory that provides a precise mathematical explanation for these experimentally observed phenomena.

\begin{figure}
\includegraphics[width=\linewidth,clip=true,trim=425 0 0 20]{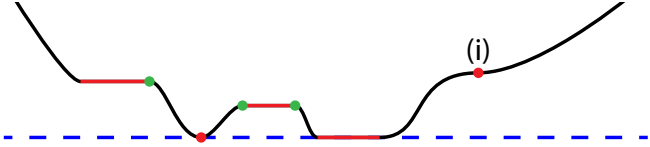}
\caption{Illustration of the properties of the framework of \cite{Haeffele:CVPR17,Haeffele:arXiv2015-PosFactor}. Starting from any initialization, a non-increasing path exists to a global minimizer. Starting from points on a plateau, a simple ``sliding" method exists to find the edge of the plateau (green points).}
\label{fig:no_local_min}
\end{figure}
%

\subsection{Optimality for neural networks with a single hidden layer}
Earlier work on global optimality for neural networks \cite{Baldi-Hornik:NN89} showed that for networks with a single hidden layer and linear activations, the squared loss has a unique global minimum and all other critical points are saddle points. However, when the activations are nonlinear, \cite{Brady:TCS99} gives examples of networks where the backpropagation algorithm \cite{werbos1974beyond} fails even with separable data. The examples are, however, not generic, and \cite{Gori-Tesi:NN91,Gori-Tesi:PAMI92} show that backpropagation generally finds a global minimizer for linearly separable data. 



Later work \cite{Bengio:NIPS2005} showed that for neural networks with a single hidden layer, if the number of neurons in the hidden layer is not fixed, but instead fit to the data through a sparsity inducing regularization, then the process of training a globally optimal neural network is analogous to selecting a finite number of hidden units from a potentially infinite dimensional space of all possible hidden units.  A weighted sum of the selected hidden units is then taken to produce the output.  The specific optimization problem is of the form
\begin{equation}
\label{eq:convex_NN_boosting}
\min_w \ell(Y,\sum_i h_i(X) w_i) + \lambda \|w\|_1,
\end{equation}
where $h_i(X)$ represents one possible hidden unit activation in response to the training data $X$ from an infinite dimensional space $h_i(X) \in \mathcal{H}$ of all possible hidden unit activations.  Clearly \eqref{eq:convex_NN_boosting} is a convex optimization problem on $w$ (assuming $\ell(Y,w)$ is convex on $w$) and straightforward to solve for a finite set of $h_i(X)$ activations.  However, since $\mathcal{H}$ is an infinite dimensional space the primary difficulty lies in how to select the appropriate hidden unit activations.  Nonetheless, by using arguments from gradient boosting, it is possible to show that problem \eqref{eq:convex_NN_boosting} can be globally optimized by sequentially adding hidden units to the network until one can no longer find a hidden unit whose addition will decrease the objective function \cite{Bengio:NIPS2005,Friedman:AnnStat2001,Mason:NIPS1999}.


\subsection{Optimality for networks with random inputs and weights}

Recent work has analyzed the problem of training neural networks with random inputs and weights. For example, the authors of \cite{Janzamin:arxiv2015} study random networks with a single hidden layer. Rather than training the network by solving an optimization problem such as 
\eqref{eq:non-convex-optimization-problem}, the authors use tensor decomposition methods to estimate the network weights from high order statistical moments of the network mapping. Moreover, the authors show that under certain assumptions on the loss and data distribution, polynomial-time training is possible \cite{Janzamin:arxiv2015}.  Further, the authors of \cite{Safran:ICML2016} study the problem when a given initialization of a neural network is likely to be within the basin of attraction of a global minimizer and provide conditions that ensure a random initialization will be within the basin of a global minimizer with high probability.

Several recent works have also analyzed the error surface of multilayer neural networks using tools from random matrix theory and statistical physics. For example, the authors of \cite{Dauphin:NIPS2014} argue that, under certain assumptions, it is vastly more likely that a critical point of a high-dimensional optimization problem be a saddle point rather than a local minimizer. Therefore, avoiding saddle points is the key difficulty in high-dimensional, non-convex optimization. Likewise, the authors of \cite{Choromanska:AIS2015} show that, under certain assumptions on the distributions of the training data and the network weight parameters, as the number of hidden units in a network increases, the distribution of local minima becomes increasingly concentrated in a small band of objective function values near the global optimum (and thus all local minima become increasingly close to being global minima).


\subsection{Global optimality for positively homogeneous networks}

Recent work \cite{Haeffele:CVPR17,Haeffele:arXiv2015-PosFactor} largely echoes ideas from the above work, but takes a markedly different approach.  Specifically, \cite{Haeffele:CVPR17,Haeffele:arXiv2015-PosFactor} analyzes the optimization problem in \eqref{eq:non-convex-optimization-problem} using a purely deterministic approach which does not make any assumptions regarding the distribution of the input data, the network weight parameter statistics, or the network initialization.  With this approach, \cite{Haeffele:CVPR17,Haeffele:arXiv2015-PosFactor} shows that saddle points and plateaus are the \textit{only} critical points that one needs to be concerned with due to the fact that for networks of sufficient size, local minima that require one to climb the objective surface to escape from, such as (f) and (h) in Fig.~\ref{fig:local_min}, are guaranteed not to exist.

More specifically, \cite{Haeffele:CVPR17} studies conditions under which the optimization landscape for the non-convex optimization problem in \eqref{eq:non-convex-optimization-problem} is such that \emph{all critical points are either global minimizers or saddle points/plateaus}, as shown in Fig.~\ref{fig:no_local_min}. The authors show that if the network size is large enough and the functions $\Phi$ and $\Theta$ are \emph{sums of positively homogeneous functions of the same degree}, any local minimizer such that \emph{some of its entries are zero} is also a global minimizer. Interestingly, ReLU and max-pooling non-linearities are positively homogeneous, while sigmoids are not, which could provide a possible explanation for the improved performance of ReLU and max pooling. Furthermore, many state-of-the-art networks are not trained with \emph{classical regularization}, such as an $\ell_1$ or $\ell_2$ norm penalty on the weight parameters but instead rely on techniques such as dropout \cite{Srivastava:JMRL2014}. The results of \cite{Haeffele:CVPR17,Haeffele:arXiv2015-PosFactor} also provide strong guidance on the design of network regularization to ensure the non-existence of spurious local minima, showing that traditional weight decay is not appropriate for deep networks.  However, more recently proposed forms of regularization such as Path-SGD \cite{Neyshabur:NIPS2015} or batch normalization \cite{Ioffe:ICML2015} can be easily incorporated into the analysis framework of \cite{Haeffele:CVPR17,Haeffele:arXiv2015-PosFactor}, and stochastic regularization methods, such as dropout \cite{Srivastava:JMRL2014}, have strong similarities to their framework.


\section{Geometric Stability in Deep Learning}
\label{sec:scattering}

An important question in the path towards understanding deep learning models is to mathematically characterize its inductive bias; i.e., define the class of regression/classification tasks for which they are predesigned to perform well, or at least better than classical alternatives. 

In the particular case of computer vision tasks, convolutional archictectures provide a fundamental inductive bias at the origin of most successful deep learning vision models. As we explain next, the notion of geometric stability provides a possible framework to understand its success.

Let $\Omega = [0,1]^d \subset \mathbb{R}^d$ be a compact $d$-dimensional Euclidean domain on which square-integrable functions $X\in L^2(\Omega)$ are defined (for example, in image analysis applications, images can be thought of as functions on the unit square $\Omega = [0,1]^2$). In a supervised learning task, an unknown function $f : L^2(\Omega) \to \mathcal{Y}$ is observed~on~a~training~set 
\begin{equation}
\{ X_i \in L^2(\Omega), Y_i = f(X_i) \}_{i \in \mathcal{I}}~,
\end{equation} 
where the target space $\mathcal{Y}$ can be thought as being discrete in a standard classification setup (with $C = | \mathcal{Y} |$ being the number of classes), or $\mathcal{Y}=\mathbb{R}^C$ in a regression task. 

In the vast majority of computer vision and speech analysis tasks, the unknown function $f$ typically satisfies the following crucial assumptions: 

1) {\bf{\em Stationarity:}} 
Consider a {\em translation operator}\footnote{
We assume periodic boundary conditions to 
ensure that the operation is well-defined over $L^2(\Omega)$.}
\begin{equation}{\cal T}_v X(u) = X(u - v), \hspace{3mm} u, v \in \Omega, 
\end{equation}
acting on functions $X \in L^2(\Omega)$. 
%
Depending on the task, we assume that the function $f$ is either {\em invariant} or {\em equivariant} with respect to translations. In the former case, we have $f({\cal T}_v X) = f(X)$ for any $X \in L^2(\Omega)$ and $v\in \Omega$. This is typically the case in object classification tasks. In the latter, we have $f({\cal T}_v X) = {\cal T}_v f(X)$, which is well-defined when the output of the model is a space in which translations can act upon (for example, in problems of object localization, semantic segmentation, or motion estimation). Our definition of invariance should not be confused with the traditional notion of \emph{translation invariant systems} in signal processing, which corresponds to translation equivariance in our language (since the output translates whenever the input translates).

2) {\bf{\em Local deformations and scale separation}:} 
Similarly, a deformation ${\cal L}_\tau$, where $\tau: \Omega \to \Omega$ is a smooth vector field,  acts on $L^2(\Omega)$ as ${\cal L}_\tau X(u) = X(u -\tau(u))$. Deformations can model local translations, changes in viewpoint, rotations and frequency transpositions \cite{bruna2013invariant}. Most tasks studied in computer vision are not only translation invariant/equivariant, but, more importantly, also stable with respect to local deformations \cite{mallat2016understanding,bruna2013invariant}. In tasks that are translation invariant we have
\begin{equation}
\label{deformstab}
| f( {\cal L}_\tau X) - f( X) | \approx \| \nabla \tau \|~,
\end{equation}
for all $X, \tau$,  where $\| \nabla \tau \|$ measures the smoothness of a given deformation field. In other words, the quantity to be predicted does not change much if the input image is slightly deformed. 
In tasks that are translation equivariant, we have 
 \begin{equation}
 \label{deformstab2}
 | f( {\cal L}_\tau X) - {\cal L}_\tau f( X) | \approx \| \nabla \tau \|.
 \end{equation}
This property is much stronger than stationarity, since the space of local deformations has high dimensionality -- of the order of $\mathbb{R}^D$ when we discretize images with $D$ pixels, as opposed to the $d$-dimensional translation group, which has only $d=2$ dimensions in the case of images. 

Assumptions \eqref{deformstab}--\eqref{deformstab2} can be leveraged to approximate $f$ from features $\Phi(X)$ that progressively reduce the spatial resolution. Indeed, extracting, demodulating and downsampling localized filter responses creates local summaries that are insensitive to local translations, but this loss of loss of sensitivity does not affect our ability to approximate $f$, thanks to \eqref{deformstab}--\eqref{deformstab2}. 
 To illustrate this principle,
denote by 
 \begin{equation}
Z(a_1, a_2;v) = \mathrm{Prob}( X(u) = a_1 \,\,\, \mathrm{and} \,\,\, X(u+v)=a_2)
 \end{equation}
 the joint distribution of two image pixels at an offset $v$ from each other. In the presence of long-range statistical dependencies, this joint distribution will not be separable for any $v$. However, the deformation stability prior states that $Z(a_1, a_2; v) \approx Z(a_1, a_2; v(1+\epsilon))$ for small $\epsilon$. 
 In other words, whereas long-range dependencies indeed exist in natural images and are critical to object recognition, they can be captured and down-sampled at different scales.
Although this principle of stability to local deformations has been exploited in the computer vision community in models other than CNNs, for instance, deformable parts models \cite{felzenszwalb2010object}, CNNs strike a good balance in terms of approximation power, optimization, and invariance. 


Indeed, stationarity and stability to local translations are both leveraged in 
convolutional neural networks (CNN).
A CNN consists of several {\em convolutional layers} of the form $\tilde{X} = C_W(X)$, acting on a $p$-dimensional input $X(u) = (X_1(u), \hdots, X_p(u))$ by applying a bank of filters $W = (w_{l,l'})$, $l = 1,\hdots, q, l' = 1,\hdots, p$ and point-wise non-linearity $\psi$, 
\begin{eqnarray}
\label{eq:convlayer}
 \tilde{X}_l(u) &=& \psi\left( \sum_{l'=1}^{p} (X_{l'} \star w_{l,l'}) (u) \right), 
\end{eqnarray}
producing a $q$-dimensional output $\tilde{X}(u) = (\tilde{X}_1(u), \hdots, \tilde{X}_q(u))$ often referred to as the {\em feature maps}.  
Here, 
\begin{equation}
\label{eq:convdef}
(X \star w)(u) = \int_\Omega  X( u-u') w(u')  du'
\end{equation}
denotes the standard convolution. 
According to the local deformation prior,
the filters $W$ 
have compact spatial support. 

Additionally, a downsampling or \emph{pooling} layer $\tilde{X} = P(X)$ may be used, defined as 
\begin{equation}
\label{eq:poolinglayer}
\tilde{X}_l(u) = P( \{ X_l( u') \, : \, u' \in \mathcal{N}(u) \} ),  \,\,\, l = 1,\hdots, q, 
\end{equation}
where $\mathcal{N}(u)\subset\Omega$ is a neighborhood around $u$ 
and $P$ is a permutation-invariant function such as an {\em average-}, {\em energy-}, or {\em max-pooling}). 

A convolutional network is constructed by composing several 
convolutional and optionally pooling layers, 
obtaining a generic hierarchical representation 
\begin{equation}
\Phi_{\boldsymbol{W}}(X) =  (C_{W^{(K)}} \cdots P \cdots \circ C_{W^{(2)}} \circ C_{W^{(1)}} ) (X),
\end{equation}
where $\boldsymbol{W} = \{W^{(1)}, \hdots, W^{(K)}\}$ is the hyper-vector of the network parameters (all the filter coefficients). 
%
The output features enjoy translation invariance/covariance depending on 
whether spatial resolution is progressively lost by means of pooling or kept fixed. 
Moreover, if one specifies the convolutional tensors to be complex wavelet decomposition operators and uses complex modulus as point-wise nonlinearities, one can provably obtain stability to local deformations \cite{mallat2012group}. Although this 
stability is not rigorously proved for generic compactly supported convolutional tensors, it underpins the empirical success of CNN architectures across a variety of computer vision applications \cite{lecun2015deep}.


A key advantage of CNNs explaining their success in numerous tasks is that the geometric priors on which CNNs are based result in a sample complexity that avoids the curse of dimensionality. Thanks to the stationarity and local deformation priors, the linear operators at each layer have a constant number of parameters, independent of the input size $D$ (number of pixels in an image). Moreover, thanks to the multiscale hierarchical property, the number of layers grows at a rate $\mathcal{O}(\log D)$, resulting in a total learning complexity of $\mathcal{O}(\log D)$ parameters. 

Finally, recently there has been an effort to extend the geometric stability priors to data that is not defined over an Euclidean domain, where the translation group is generally not defined. In particular, researchers are exploiting geometry in general graphs via the spectrum of graph Laplacians and its spatial counterparts; see \cite{bronstein2016geometric} for a recent survey on those advances.

\section{Structure Based Theory for Deep Learning}
\label{sec:structure-generalization}

\subsection{Structure of the data throughout a neural network}
\label{sec:structure}

An important aspect for understanding better deep learning is the relationship between the structure of the data and the deep network. For a formal analysis, consider the case of a network that has random i.i.d. Gaussian weights, which is a common initialization in  training deep networks. Recent work \cite{Giryes2015} shows that such networks with random weights preserve the metric structure of the data as they propagate along the layers, allowing for stable recovery of the original data from the features calculated by the network -- a property that is often encountered in general deep networks \cite{Bruna14Signal, Mahendran15Understanding}. 

More precisely, the work of \cite{Giryes2015} shows that the input to the network can be recovered from the network's features at a certain layer if their size is proportional to the intrinsic dimension of the input data. This is similar to data reconstruction from a small number of random projections \cite{Candes06Near, Chandrasekaran12Convex}. However, while random projections preserve the Euclidean distance between two inputs up to a small distortion, each layer of a deep network with random weights distorts this distance proportionally to the angle between the two inputs: the smaller the angle, the stronger the shrinkage of the distance. Therefore, the deeper the network, the stronger the shrinkage achieved. Note that this does not contradict the fact that it is possible to recover the input from the output; even when properties such as lighting, pose and location are removed from an image (up to certain extent), the resemblance to the original image is still maintained.

As random projection is a universal sampling strategy for low-dimensional data \cite{Candes06Near,Chandrasekaran12Convex, Giryes2017}, deep networks with random weights are a universal system that separates any data (belonging to a low-dimensional model) according to the angles between the data points, where the general assumption is that there are large angles between different classes \cite{Wolf03Learning, Elhamifar:TPAMI13}. As the training of the projection matrix adapts it to better preserve specific distances over others, the training of a network prioritizes intra-class angles over inter-class ones. This relation is alluded to by the proof techniques in \cite{Giryes2015} and is empirically manifested by observing the angles and Euclidean distances at the output of trained networks.

By using the theory of 1-bit compressed sensing, \cite{Plan14Dimension} shows that each layer of a network preserves the metric of its input data in the Gromov-Hausdorff sense up to a small constant $\delta$, under the assumption that these data reside in a low-dimensional manifold denoted by $K$. This allows drawing conclusions on the tessellation of the space created by each layer and the relationship between the operation of these layers and local sensitive hashing (LSH). It also implies that it is possible to retrieve the input of a layer, up to certain accuracy, from its output. This shows that every layer preserves the important information of the data.

An analysis of the behavior of the Euclidean distances and angles in the data along the network reveals an important effect of the ReLU. Without a non-linearity, each layer is simply a random projection for which Euclidean distances are approximately preserved. The addition of a ReLU makes the system sensitive to the angles between points. In this case, the network tends to decrease the Euclidean distances between points with small angles between them (``same class''), more than the distances between points with large angles between them (``different classes''). Still, low-dimensional data at the input remain such throughout the entire network, i.e., deep networks (almost) do not increase the intrinsic dimension of the data \cite{Giryes2015}. This is related to the recent work in \cite{Papyan16} that claims that deep networks may be viewed a sparse coding procedure leading to guarantees on the uniqueness of the representation calculated by the network and its stability. 

As random networks are blind to the data labels, training may select discriminatively the angles that cause the distance deformation. Therefore, it will cause distances between different classes to increase more than the distances within the same class. This is demonstrated in several simulations for different deep networks. It is  observed that a potential main goal of the training of the network is to treat the class boundary points while keeping the other distances approximately the same.

\subsection{Generalization error}
\label{sec:generalization}

The above suggests that there is a relation between the structure of the data and the error the network achieves in training, which leads to study the relationship between the generalization error in deep networks and the data structure. Generalization error -- the difference between the empirical error and the expected error -- is a fundamental concept in statistical learning theory. Generalization error bounds offer insight into why learning from training samples is possible. 

Consider a classification problem with a data point $X \in \mathcal{X} \subseteq \mathbb{R}^{D}$ that has a corresponding class label $Y \in \mathcal{Y}$, where $C$ is the number of classes. A training set of $N$ samples drawn from a distribution $P$ is denoted by $\Upsilon_N =\{(X_i, Y_i) \}_{i=1}^N$ and the loss function is denoted by $\ell(Y, \Phi(X,W))$, which measures the discrepancy between the true label $Y$ and the estimated label $\Phi(X,W)$ provided by the classifier. 
The  empirical loss of a network $\Phi(\cdot,W)$ associated with the training set $\Upsilon_N$ is defined as
\begin{eqnarray}
	\ell_{\text{emp}}(\Phi) = \frac{1}{N} \sum_{X_i \in \Upsilon_N} \ell \left(Y_i,\Phi(X_i, W)\right),
\end{eqnarray}
and the expected loss as 
\begin{eqnarray}
	\ell_{\text{exp}}(\Phi) = \mathbb{E}_{(X,Y) \sim P} \left[ \ell\left(Y, \Phi(X,W) \right) \right].
\end{eqnarray}
The \textit{generalization error} is then given as:
\begin{eqnarray}
	\text{GE}(\Phi) = |\ell_{\text{exp}}(\Phi) - \ell_{\text{emp}}(\Phi)| \,.
\end{eqnarray}

Various measures such as the the VC-dimension \cite{Vapnik1999,Shalev-Shwartz2014}, the Rademacher or Gaussian complexities \cite{Bartlett2003} and algorithm robustness \cite{Xu2012a} have been used to bound the GE in the context of deep networks. However, these measures do not offer an explanation for good deep network generalization in practice where the number of parameters can often be larger than the number of training samples \cite{Zhang2017} or the networks are very deep \cite{He2015}. For example, the VC-dimension of a deep network with the hard-threshold non-linearity is equal to the number of parameters in the network, which implies that the sample complexity is linear in the number of parameters of the network. On the other hand, the work \cite{Neyshabur2015} bounds the generalization error independently of the number of parameters. Yet, in its bound the generalization error  of deep network with ReLUs scales exponentially with the network depth. Similarly, the authors in \cite{Xu2012a} show that deep networks are robust provided that the $\ell_1$-norm of the weights in each layer is bounded. The bounds are exponential in the $\ell_1$-norm of the weights if the norm is greater than 1. 

An alternative route followed by \cite{Sokolic2016, Sokolic2017b, Xu2012a, Huang15Discriminative} bounds the generalization error in terms of the networks' classification margin, which is independent of  the depth and size but takes into account the structure of the data (considering its covering number) and therefore avoids the above issues. As it is hard to calculate the input margin directly, in \cite{Sokolic2016} it is tied to the Jacobian matrix and the loss of the deep networks showing that bounding the spectral norm of the Jacobian matrix reduces the generalization error. This analysis is general to arbitrary network structures, types of non-linearities and pooling operations. Furthermore, a relationship between the generalization error, the invariance in the data and the network is formally characterized in \cite{Sokolic2017b}.

\begin{table}[t]
\centering
\caption{Classification acc. $[\%]$ of ResNet CIFAR-10} \label{tab:resnet_results}
\begin{tabular}{ccc}
    \toprule
\# train samples & ResNet & ResNet + Jac. reg.  \\
\midrule
2500 & 55.69  & \textbf{62.79 } \\
10000  &  71.79 & \textbf{78.70} \\
50000 + aug. & 93.34 & \textbf{94.32} \\
 \bottomrule
\end{tabular}
\vspace{-1mm}
\end{table}

Using the relationship between the generalization error and the network Jacobian matrix, a new Jacobian based regularization strategy is developed in \cite{Sokolic2017b} and its advantage is demonstrated for several networks and datasets. For example, Table~\ref{tab:resnet_results} shows the improvement achieved when using this regularization with the Wide ResNet architecture \cite{Zagoruyko2016} for  CIFAR-10 with different numbers of training examples.

A related theory to the one presented above is the one in \cite{Neyshabur:NIPS2015, Hardt:ICML16}. These works study the relationship between the generalization error and the minimization of the network loss using SGD. They provide modifications for SGD that improves the error achieved by the network. 

\section{Towards an Information-Theoretic Framework}
\label{sec:info-theory}

The loss function of choice 
for  training deep networks to solve supervised classification problems is the {\em empirical cross-entropy}
\begin{equation}
\tilde \ell(W) = {\mathbb E}_{P(X,Y)}(-\log \Phi(X,W)),
\end{equation}
This loss function is prone to overfitting, as the network could trivially memorize the training data instead of learning
the underlying distribution.
This problem is usually addressed by regularization, which can be explicit (e.g., the norm of $W$, known as \emph{weight decay}), or implicit in stochastic gradient descent. It was suggested by \cite{hinton1993keeping} almost a quarter century ago that better regularization, hence less overfitting, might be achieved by limiting the information stored in the weight, $\KL{p(W|X,Y)}{p(W)}$, where $p(W)$ is a prior on the weights. Choosing this regularizer leads to the loss function
\begin{equation}
\label{eq:weight-ib}
\ell(W) = {H}(Y | X, W) + \lambda \KL{p(W|X,Y)}{p(W)}
\end{equation}
where the first term denotes the empirical conditional cross-entropy obtained from $\tilde \ell(W)$. For $\lambda=1$, this is the variational lower bound on the observed data distribution $p_\theta(Y|X)$, and can therefore be seen as a form of {\bf Bayesian inference} of the weights. More generally, this is equivalent to the information bottleneck Lagrangian. The first term is the same as the empirical cross-entropy and ensures that information stored in the weights is \emph{sufficient} for the task $Y$, while the second term minimizes the amount of information stored. Thus, the weights learned by minimizing cross-entropy, with a KL regularizer, approximate a minimal sufficient statistic of the training set.
Computing the KL term and optimizing $\ell$ was considered a show-stopper until recently, when advances in Stochastic Gradient Variational Bayes \cite{kingma2015variational,kingma2013auto} made efficient optimization possible. 



But for a representation to be useful, it should not just efficiently memorize past (training) data. It should also reduce the effect of nuisance variability affecting future (test) data. Indeed, most of the variability in imaging data can be ascribed to the effects of illumination and viewpoint, quotienting out which leaves a thin set \cite{sundaramoorthiPVS09}. As we have already pointed out, deep networks are known to reduce the effects of nuisance variability in test data. This can be partly achieved through the architecture, in particular multi-scale convolutional structures. Some may be ascribed to the optimization procedure, that converges to ``flat minima.'' But the choice of regularizer is also responsible for the networks' ability to discount nuisance variability. Denoting by $X$ be the input sample, $Y$ the target variable, and $Z \sim p(Z|X)$ the (stochastic) representation of $X$ learned by a layer in the network, the tradeoff between sufficiency an minimality of $Z$ is formalized by the information bottleneck Lagrangian
\begin{equation}
\label{eq:representation-ib}
\ell(W) = H(Y\,|\, Z, W) + \lambda I(Z;X),
\end{equation}
where the first term ensures that the representation $Z$ is sufficient for $Z$, while the second term ensures its information content remains minimal, i.e.\ that nuisances are filtered out. 

Notice that, while formally equivalent, the losses in (\ref{eq:weight-ib}) and (\ref{eq:representation-ib}) are  conceptually different: In the former, the weights are a representation of the \emph{training set} that is minimal and sufficient. In the latter, the activations are a minimal representation of the \emph{test sample}. We conjecture that the two are related, and that the relation informs the generalization properties of the network, but specific bounds have yet been shown.

Different choices of the noise distribution, and different models of the marginal distribution $p(Z)$ lead to slightly different analyses. For example, \cite{alemi2016deep} considers the case of additive Gaussian noise and Gaussian marginals, while \cite{achille2016information} study multiplicative noise distributions, and considers both a scale invariant log-uniform marginal (the only one compatible with the fact that networks with ReLU activations are scale invariant), and a log-normal marginal distribution. Interestingly, the special case in which the multiplicative noise is chosen from a Bernoulli distribution reduces to the well-known practice of Dropout \cite{hinton1993keeping}, while choosing from a multiplicative Gaussian distribution leads to Gaussian Dropout \cite{kingma2013auto}.

In order to compute the term $I(Z;X)$, it is commonly assumed that the activations are mutually independent, that is, that the marginal $p(Z)$ is factorized; \cite{achille2016information} shows that making this assumption corresponds to minimizing a modified loss function, which also reduces the total correlation $\operatorname{TC}(Z)$ of the activations. Therefore, a choice dictated by convenience in order to explicitly compute the information bottleneck Lagrangian, yields a disentangled representation, with entanglement measured by total correlation.

It is remarkable that empirical practice has managed to converge to the use of the cross-entropy loss with dropout, that happens to be what would have been prescribed by first principles, since for certain choices of distribution, training is equivalent to minimizing the information bottleneck Lagrangian, that yields an approximation of a minimal sufficient invariant statistic, which define an optimal representation. It is only recently that developments in theory have made this association possible, and developments in optimization and hardware have made deep neural networks a sensational success.




\section{ACKNOWLEDGMENTS}

Ren\'e Vidal acknowledges grant NSF 1618485.
Raja Giryes acknowledges the Global Innovation Fund (GIF).
Stefano Soatto acknowledges grants ONR N00014-15-1-2261, ARO W911NF-15-1-0564/66731-CS, and AFOSR FA9550-15-1-0229.

\bibliographystyle{unsrt}
\bibliography{CDC17-Tutorial,bibliography}

\end{document}